\documentclass[10pt,twocolumn,letterpaper]{article}

\usepackage{iccv}
\usepackage{times}
\usepackage{epsfig}
\usepackage{graphicx}
\usepackage{amsmath}
\usepackage{amssymb}
\usepackage{amsfonts}
\usepackage{color}
\usepackage{colortbl}
\usepackage{epsfig}
\usepackage{multirow}
\usepackage{algorithm}
\usepackage{subcaption}
\usepackage{algpseudocode}
\usepackage{scrextend}
\usepackage{tabularx}
\usepackage{booktabs}
\usepackage{bbm}
\usepackage{breqn}
\pdfoutput=1
\newcommand{\vect}[1]{\boldsymbol{#1}}
\def\smallint{\begingroup\textstyle \int\endgroup}

\definecolor{tabrow}{rgb}{0.88,0.88,0.88}
\definecolor{Gray}{gray}{0.8}

\newcommand{\dc}{2DC\xspace}

\newcommand{\dccnn}{DC$_{p}$\xspace}
\newcommand{\vgStef}{VO-GPAE\xspace}
\newcommand{\cnn}{CNN \cite{gudi2015deep}\xspace}
\newcommand{\vgg}{VGG16 \cite{Simonyan14c}\xspace}
\newcommand{\ocnn}{OR-CNN \cite{niu2016ordinal}\xspace}
\newcommand{\dcod}{{\it DeepCoder}\xspace}
\newcommand{\drcnn}{CCNN-IT \cite{walecki2017deep}\xspace}

\usepackage[pagebackref=true,breaklinks=true,letterpaper=true,colorlinks,bookmarks=false]{hyperref}

\iccvfinalcopy 

\def\iccvPaperID{1409} 

\ificcvfinal\fi
\begin{document}

\title{DeepCoder: Semi-parametric Variational Autoencoders \\for Automatic Facial Action Coding}

\author{Dieu Linh Tran\thanks{These authors contributed equally to this work.} , Robert Walecki, Ognjen (Oggi) Rudovic*, Stefanos Eleftheriadis, \\
Bj{\"o}rn Schuller and Maja Pantic\\
{\tt\small \{linh.tran, r.walecki14, bjoern.schuller, m.pantic\}@imperial.ac.uk}\\
{\tt\small stefanos@prowler.io}\\
{\tt\small orudovic@mit.edu}
}




\maketitle

\begin{abstract}
Human face exhibits an inherent hierarchy in its representations (i.e., holistic facial expressions can be encoded via a set of facial action units (AUs) and their intensity). Variational (deep) auto-encoders (VAE) have shown great results in unsupervised extraction of hierarchical latent representations from large amounts of image data, while being robust to noise and other undesired artifacts. Potentially, this makes VAEs a suitable approach for learning facial features for AU intensity estimation. Yet, most existing VAE-based methods apply classifiers learned separately from the encoded features. By contrast, the non-parametric (probabilistic) approaches, such as Gaussian Processes (GPs), typically outperform their parametric counterparts, but cannot deal easily with large amounts of data. To this end, we propose a novel VAE semi-parametric modeling framework, named {\dcod}, which combines the modeling power of parametric (convolutional) and non-parametric (ordinal GPs) VAEs, for joint learning of (1) latent representations at multiple levels in a task hierarchy\footnote{The benefit of using VAE for hierarchical learning of image features in an unsupervised fashion has been shown in \cite{masci2011stacked}, which is particularly important for addressing the hierarchy in face representation: low - sign level (AUs), high - judgment level (emotion expressions) \cite{ekman2002facial}.}, and (2) classification of multiple ordinal outputs. We show on benchmark datasets for AU intensity estimation that the proposed \dcod outperforms the state-of-the-art approaches, and related VAEs and deep learning models.
\end{abstract}

\section{Introduction}
Automated analysis of facial expressions has many applications in health, entertainment, marketing and robotics, where measuring facial affect can help to make inferences about the patient's conditions, user's preferences, but also enable more user-friendly and engaging technology. Facial expressions are typically described in terms of the configuration and intensity of facial muscle actions using the Facial Action Coding System (FACS) \cite{ekman2002facial}. FACS defines a unique set of 30+ atomic non-overlapping facial muscle actions named Action Units (AUs) \cite{kohn2009cvpr}, with rules for scoring their intensity on a six-point ordinal scale. Using FACS, nearly any anatomically possible facial expression can be described as a combination of AUs and their intensities.
\begin{figure}
\includegraphics[width=0.5\textwidth]{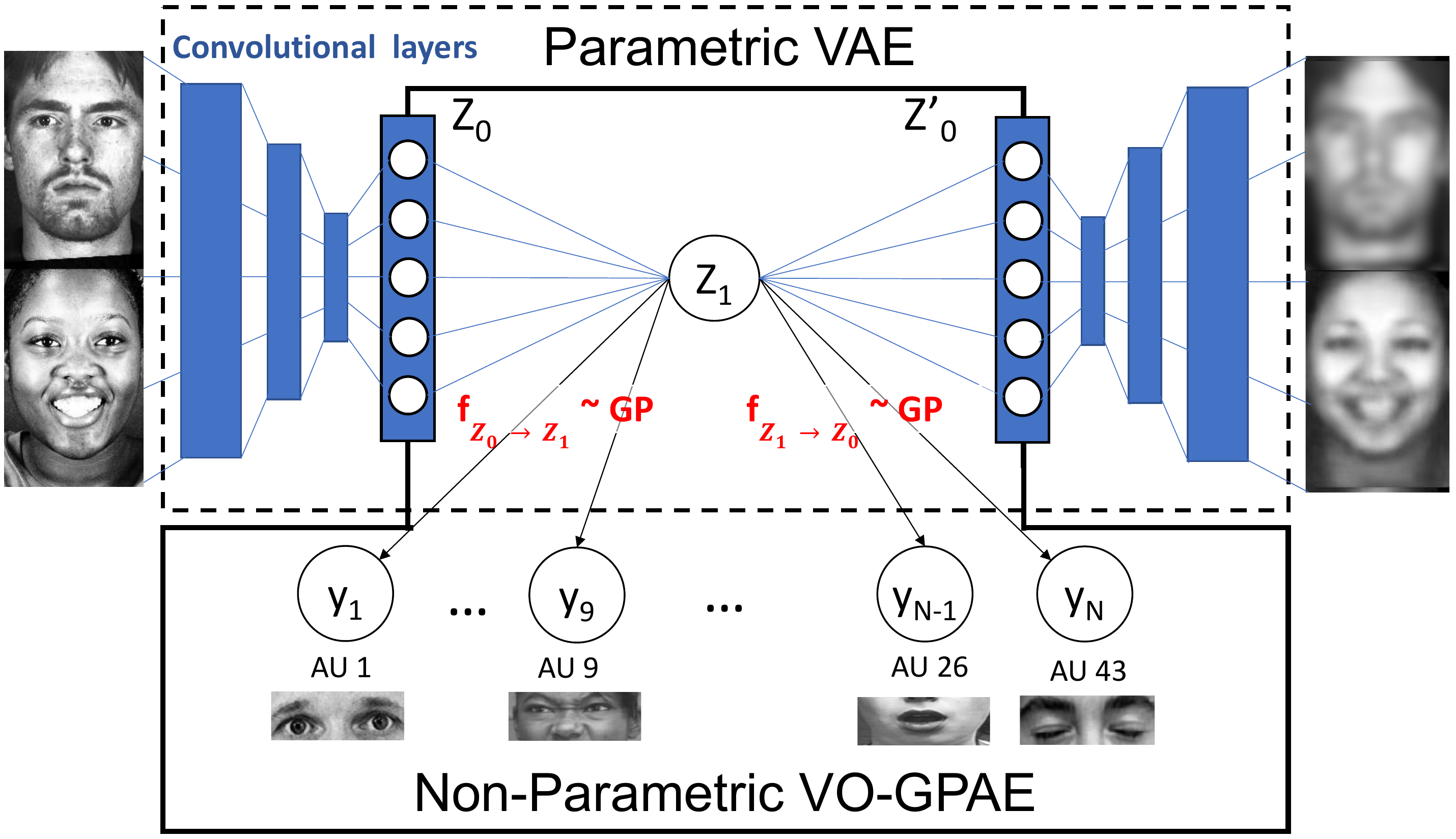}
\caption{The proposed 2-layer \dcod: the input is a face image, and the outputs are the  reconstructed face image and AU intensity levels. The top variational convolutional autoencoder (VAE) performs the first level coding ($\vect{Z}_0$) of the facial features, while further encoding ($\vect{Z}_1$) of these features is optimized for AU intensity estimation using ordinal GP variational autoencoder (VO-GPAE).}\label{model_intro}
\end{figure}
However, despite the rapid growth in available facial images (videos), there is an overall lack of annotated images (in terms of AUs). This is mainly because it entails a costly and time-consuming labeling effort by trained human annotators. For instance, it may take more than an hour for the expert annotator to code the intensity of AUs in one second of a face video. Even then, the annotations are biased, resulting in a low agreement between the annotators. This is further challenged by a large variability in imaging conditions, facial morphology and dynamics of expressions. Therefore, there is a need for machine learning models that can efficiently and accurately perform the AU coding of target face images.

Recent advances in deep neural networks (DNN), and, in particular, convolutional models (CNNs) \cite{gudi2015deep}, have shown great advances towards automating the process of image coding. The effectiveness of these models has been demonstrated on many general vision problems \cite{krizhevsky2012imagenet,szegedy2013deep,sun2014deep}. In the context of facial expression analysis, the majority of existing 'deep' works  consider only baseline tasks such as expression recognition and AU detection \cite{liu2013aware,zhao2016deep,khorrami2015deep}. Only a handful of these works attempted AU intensity estimation \cite{gudi2015deep}. This is due to the limited annotated face images of AU intensity (that otherwise could fully be exploited in deep learning), and the difficulty in discerning AU intensities. 

Traditionally, the AU intensity estimation has been addressed by non-deep models (SVMs, CRFs, etc.) \cite{walecki2016copula,kaltwang2015}, and using geometric features such as the locations of characteristic facial points, and/or hand-crafted appearance-based features (such as LBPs, Gabors or SIFT). An alternative approach that is being commonly adopted in a variety of computer vision tasks is to automatically extract most informative features from (high-dimensional) input images using the notion of convolutitonal auto-encoders (CAE) \cite{masci2011stacked,van2016conditional,kipf2016variational}. CAE  differ from conventional AEs~\cite{ciregan2012multi} as they are built using convolutional layers with shared weights among neighborhood pixels that preserve the spatial locality. The CAE architectures are typically similar to that of a CNN with additional inverse convolution operation~\cite{masci2011stacked}. The key ingredient of CAEs is that they are learned by minimizing the reconstruction loss without the need for image labels, while reducing the effects of noise in the input. 

Consider a practical example typically occurring in automated analysis of facial expressions, and, in particular, AU intensity coding: we have access to a large corpora of unlabeled face images, but only a few thousand images are coded in terms of AU intensity. To fully leverage the available data, efficient and highly expressive generative models based on VAE can be used to find a set of underlying features from unlabeled images. Due to the reconstruction cost of VAEs, it is assured that the obtained features represent well the high dimensional face images. Then, highly expressive non-parametric prediction models (e.g., based on GPs~\cite{rasmussen2006gaussian}) can be applied. This allows them to focus on the main task - in our case, the AU intensity estimation, instead of the computationally expensive feature selection. More importantly, such non-parametric approaches when applied to robust input features are expected to generalize better than their parametric counterparts (e.g., soft-max output layer of DNNs) due to the ability to preserve specific structures in target features -- such as subject-specific variation in AU intensity. This is achieved by means of their kernel functions that can focus on data samples in the VAE feature space, effectively doing smoothing over training subjects to make best prediction of AU intensity levels for the test subject. 

While the approach described above is a promising avenue for the design of a class of semi-parametric auto-encoding models, independently applying the two models (e.g., VAE for feature extraction, and non-parametric models for AU intensity estimation) is suboptimal as there is no sharing of information (and parameters). To this end, we propose a novel model, named \dcod, that leverages the power of parametric and non-parametric VAEs in a unified probabilistic framework. Specifically, \dcod is a general framework that builds upon a hierarchy of any number of VAEs, where each coding/decoding part of the intermediate VAEs interacts with the neighboring VAEs during learning, assuring the sharing of information in both directions (bottom-up \& top-down). This is achieved through a newly introduced approximate learning of VAEs in \dcod. We illustrate this approach by designing an instance of \dcod as a two-level semi-parametric VAE (\dc) - the top level being the standard parametric VAE \cite{kingma2013auto}, and the bottom level (also used for AU intensity estimation) being a non-parametric Variational Ordinal Gaussian Process AE (\vgStef) \cite{eleftheriadis2016variational}. We choose these two approaches as their probabilistic formulation allows for tying of their priors over the latent features, in a principled manner. The model is depicted in Fig.~(\ref{model_intro}). We show on two benchmark datasets for AU intensity estimation (DISFA\cite{mavadati2013disfa} and FERA \cite{valstar2015fera}) that the proposed approach outperforms the state-of-the-art approaches for the AU intensity estimation.

\section{Related Work}
\subsection{Facial Action Unit Intensity Estimation}
Estimation of AUs intensity is often posed as a multi-class problem approached using Neural Networks \cite{kapoor2005multimodal}, Adaboost \cite{bartlett2006automatic}, SVMs \cite{lucey2007investigating} and belief networks \cite{liu2014facial} classifiers. Yet, these methods are limited to a single output, thus, a separate classifier is learned for each AU - ignoring the AU dependencies. This has been addressed using the multi-output learning approaches. For example, \cite{nicolle2015facial} proposed a multi-task learning for AU detection, where a metric with shared properties among multiple AUs was learned. Similarly, \cite{sandbach2013markov} proposed a MRF-tree-like model for joint intensity estimation of AUs. \cite{kaltwang2015} proposed Latent-Trees (LTs) for joint AU-intensity estimation that captures higher-order dependencies among the input features and AU intensities. More recently, \cite{walecki2016copula} proposed a multi-output Copula Regression for ordinal estimation of AU intensity. However, these cannot directly handle high-dimensional input face images.

\subsection{CNNs for Facial Expression Analysis}
CNNs operate directly on the input face images to extract optimal image features. \cite{liu2013aware} introduced an AU-aware receptive field layer in a deep network, designed to search subsets of the over-complete representation, each of which aims at simulating the best combination of AUs. Its output is then passed through additional layers aimed at the expression classification, showing a large improvement over the traditional hand-crafted features. In \cite{gudi2015deep}, a CNN is jointly trained for detection and intensity estimation of AUs. More recently, \cite{zhao2016deep} introduced an intermediate region layer learning region specific weights. These methods are parametric, with the CNN used to extract deep features; yet, the network output remains unstructured. Thus, none of these models exploits CNNs in the context of (ordinal) deep semi-parametric models, as done in \dcod. Note also that in \dcod we exploit a label-augmented version of VAEs, which can be seen as a variant of CNNs used for classification, but with an additional noise-reduction cost (decoder).

\subsection{Autoencoders (AE)}
The main idea of AEs is to learn latent representations automatically from inputs, usually in an unsupervised  manner \cite{masci2011stacked, baldi2012autoencoders, le2013building}. Recenly, variational AEs (VAEs) have gained attention as parametric generative models \cite{gregor2015draw, kingma2013auto, kingma2014semi, sohn2015learning} and their stacked or convolutional variations \cite{kulkarni2015deep, larsen2015autoencoding}. Example applications include the reconstruction of noisy and/or partially missing data \cite{vincent2008extracting, vincent2010stacked}, or feature extraction for classification \cite{ciregan2012multi}. Furthermore, AEs based on deep networks have shown their efficacy in many face-related recognition problems \cite{kan2014stacked, liu2014facial, zhang2013occlusion}.

AEs are also closely related to GP Latent Variable Models (GPLVMs) with "back-constraints"~\cite{lawrence2005probabilistic,titsias2010bayesian,snoek2012nonparametric}. This mapping facilitates a fast inference mechanism and enforces structure preservation in the latent space. In~\cite{damianou2015semi,dai2015variational}, the authors proposed a variational approximation to the latent space posterior. \cite{hensman2014nested} proposed deep GPs for unsupervised data compression. More recently,~\cite{eleftheriadis2016variational} introduced a Variational Ordinal GP AE (\vgStef), which includes a GP mapping as the decoding model. This allows \vgStef to learn the GP encoders/decoders in a joint framework. We extend this formulation of the non-parametric VAE by embedding it into the bottom layer of \dcod, while using the (parametric) convolutional AE at the top - achieving an efficient feature extraction. 

\section{DeepCoder: Methodology}

Assume we are given a training dataset $\mathcal{D} = \lbrace \vect{X}, \vect{Y}\rbrace$, with $N_{\mathcal{D}}$ input images $\vect{X} = [\vect{x}_1, \ldots, \vect{x}_i, \ldots, \vect{x}_{N_{\mathcal{D}}}]^T$. The corresponding labels $\vect{Y} = [\vect{y}_1, \ldots, \vect{y}_i, \ldots, \vect{y}_{N_{\mathcal{D}}}]^T$ are comprised of multivariate outputs stored in $\vect{y}_i = \lbrace \vect{y}^{1}_i, \ldots \vect{y}^{q}_i, \ldots \vect{y}^{Q}_i\rbrace$, where $Q$ is the number of AUs, and $\vect{y}^{q}_i$ takes one of $\{1,...,{L^q}\}$ intensity levels. Our goal is to predict $\vect{y}_*$ and reconstruct $\vect{x}_{*}^'$, given a new test input image $\vect{x}_*$. To learn the highly non-linear mappings $\vect{X} \to \vect{Y}$, we perform encoding and decoding of input features $\vect{X}$ via multiple layers of VAEs. These layers are encoded by the latent variables $\vect{Z} = \{\vect{Z}_i\}, \; i = 0, \dots, N-1$,  where the dimension of $\vect{Z}_i$ can vary for each $i$, and $N$ is the number of layers. For simplicity, we first assume a single VAE layer with latent variables $\vect{Z}_0$. This leads to the following marginal log-likelihood and its corresponding variational lower bound: 
\begin{align}\label{eq:marginallog}
\log p(\vect{X}, \vect{Y}) = &\log \int p(\vect{X}|\vect{Z}_0) p(\vect{y}|\vect{Z}_0) p(\vect{Z}_0)d\vect{Z}_0\\
\begin{split}
\ge \; & \mathbb{E}_{q(\vect{Z}_0|\vect{X})} \big[ \log p(\vect{X}| \vect{Z_0})\big] \\ \label{lowerbound}
& + \mathbb{E}_{q(\vect{Z}_0|\vect{X})} \big[ \log p(\vect{Y} | \vect{Z}_0)\big]\\
& - D_{KL}(q(\vect{Z}_0|\vect{X})||p(\vect{Z_0})) \\
\end{split}
\end{align}

\begin{figure}[ht]
\begin{center}
\includegraphics[width=\linewidth]{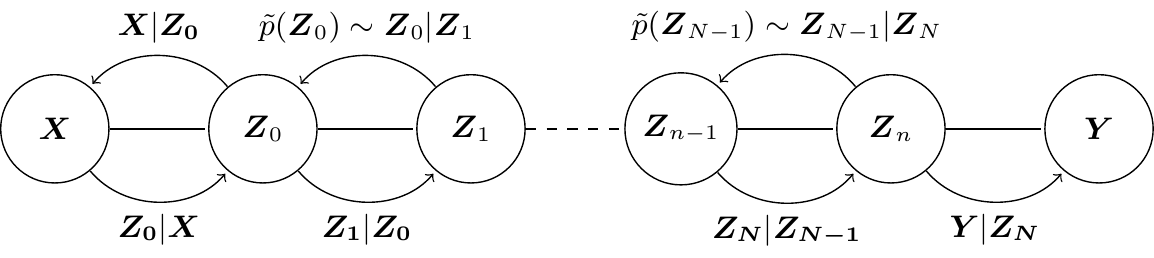}
\end{center}
   \caption{The general formulation of \dcod as a stack of $N$ VAEs, modeling the input-output pairs: ($\vect{X}$,$\vect{Y}$). The conditionals $(\vect{Z}_0|\vect{X}), \ldots (\vect{Y}|\vect{Z}_N)$ from left to right in \dcod perform the coding part, while from right to left perform the decoding part via $(\vect{Z}_{N-1}|\vect{Z}_N), \ldots (\vect{X}|\vect{Z}_0)$. Note that for N=2, we obtain the proposed 2-layer \dcod, modeled using VC-AE and VO-GPAE, respectively.}
\label{fig:model}
\label{model_graph}
\end{figure}

In Eq.~(\ref{eq:marginallog}), the first two terms are the reconstruction loss over the input features and output labels, respectively, under the estimated posterior. The second term is the Kullback-Leibler (KL) divergence which measures the difference between the approximate and true posterior. We obtain the latter by exploiting the conditional independence $\vect{X} \perp\!\!\!\perp \vect{y} | \vect{Z}_0$ (see \cite{eleftheriadis2016variational} for details). To account for more complex dependencies between $(\vect{X}, \vect{Y})$, we generalize Eq.~(\ref{lowerbound}) by expanding $p(\vect{Z}_0)$ as a stack of $N$ VAE layers (see Fig.~(\ref{model_graph}))
\begin{align}
\begin{split}
&\underbrace{\smallint p(\vect{Z}_0|\vect{Z_1}) \ldots  \underbrace{\smallint p(\vect{Z}_{N-1}|\vect{Z_{N}}) p(\vect{Z_N}) d\vect{Z}_N}_{\substack{\tilde{p}(\vect{Z}_{n-1}) \\ \vdots}} \ldots d\vect{Z}_1}_{\tilde{p}(Z_0)}
\end{split}
\end{align}
This approach has high modeling power; however, it comes with the cost of having to simultaneously learn multiple (deep) layers of latent variables $\vect{Z}$. While this is computationally tractable for a single layer ($\vect{Z}_0$), in the case of more layers, we need to resort to approximate methods. To this end, we propose an optimization approach that sequentially performs a chain-like propagation of uncertainty of each coder. Specifically, we solve for the posteriors of each coder 'locally' and use the learned posteriors to define the (approximate) prior $p(\vect{Z})$, needed to compute the KL divergence of each subsequent coder in the sequence from 'bottom-up' (a practical example of this is described in Alg.~(\ref{alg})). For the $(N-1)$-th VAE, instead of using a flat Gauss prior $p(\vect{Z}_{N-1})$, we approximate it using the posterior of the $N$-th decoder learned as:
\begin{align}
\begin{split}
\log \tilde{p}(\vect{Z}_{n-1}) \ge & \mathbb{E}_{q(\vect{Z}_{N}|\vect{Z}_{N-1})} \big[ \log p(\vect{Z}_{N-1}| \vect{Z}_N)\big] \\
&- D_{KL}(q(\vect{Z}_N|\vect{Z}_{N-1})||\tilde{p}(\vect{Z_N}))
\end{split}
\end{align}
Note the main benefit of the proposed: instead of assuming a flat  prior over the latent variables, as typically done in existing VAE \cite{kingma2013auto}, we define the priors on $\vect{Z}$ that are informed of the uncertainty of each coder 'below' in the deep structure, while also retaining the information about the decoding error of all subsequent coders. Thus, by exploiting the conditional independence of $\vect{Z}$ at each level of \dcod, we seamlessly 'encode' complex relationships between $\vect{X}$ and $\vect{Y}$. From the regularization perspective, we constrain the parameters via the KL terms (based on priors $\tilde{p}(\vect{Z})$) at each level of coding/decoding in \dcod. Fig.~(\ref{fig:model}) illustrates the main idea for the general case. Using this framework, we generate an instance of \dcod as a two-layer semi-parametric coder: the top coder takes the parametric form (Convolutional VAE) and the bottom the non-parametric form (VAE based on GPs). We choose these two because their probabilistic formulation allows us to combine them in a Bayesian framework. Also, instead of using directly CNNs in the first layer, we opt for using VAEs due to their de-noising of input features (although we augment the subspace learning using target labels as in CNNs).
\subsection{Variational Convolutional AEs (VC-AE)}
\label{vcae}
In the top layer, we use the VC-AE to map the inputs $\vect{X}$ onto the latent space $\vect{Z}_0$. A decoder network is then used to map these latent space points back to the original input data. Formally, the parameters of VC-AE are learned by maximizing the objective:
\begin{align}\label{eq:lossCVAE}
\begin{split}
\mathcal{L}_{VC-AE}(W_d, \mu, \sigma) &= \mathcal{L}_{kl, \vect{X}} + \mathcal{L}_{r, \vect{X}}\\
\mathcal{L}_{kl, \vect{X}} &=-D_{KL}(q_{\mu,\sigma}(\vect{Z}_0|\vect{X})||p(\vect{Z}_0)) \\
\mathcal{L}_{r, \vect{X}} &= \mathbb{E}_{q(\vect{Z}_0|\vect{X})} \big[ \log p(\vect{X}| \vect{Z}_0)\big]
\end{split}
\end{align}
where the KL divergence ($\mathcal{L}_{kl, \vect{X}}$) and reconstruction term ($\mathcal{L}_{r, \vect{X}}$), form the variational lower bound typically optimized in VC-AEs. The conditionals are parametrized as:
\begin{align}\label{ae}
 \vect{X}|\vect{Z}_{0} &= f_{\vect{Z}_{0} \to \vect{X}}(\cdot; \theta_{\vect{Z}_{0} \to \vect{X}}), \\ 
 \vect{Z}_{0}| \vect{X} &= f_{\vect{X} \to \vect{Z}_0}(\cdot;\theta_{\vect{X} \to \vect{Z}_0}).
\end{align}
Their functional forms are given by the VC encoder ($ \vect{Z}_{0}| \vect{X}$) and decoder ($\vect{X}|\vect{Z}_{0})$. For the convolutional coder part ($ \theta_{\vect{Z}_{0} \to \vect{X}}$),  we used 5 convolutional layers containing 128, 64, 32, 16 and 8 filters. The filter size was set to $5\times5$ pixel followed by ReLu (Rectified Linear Unit) activation functions \cite{kingma2013auto}. We also used $2\times2$ max pooling layers after each convolutinonal layer. The compressed representations are $15 \times 20 \times 16$ pixels and are passed to two fully connected layers, which return $2000$ features each, with the latent space variational posterior $q(\vect{Z}_0|\vect{X})\sim \mathcal{N}(\mu, \sigma^2)$. For deconvolution ($\theta_{\vect{X} \to \vect{Z}_0}$), we used up-scaling instead of max-pooling and deployed the inverse encoder architecture. For this, we exploited the re-parameterization trick \cite{kingma2013auto}. We sample points $z$ at random from the distribution of latent variables $\vect{Z}_0$, in order to generate the data. Finally, the decoder network maps $z$ back to the original input.
\subsection{Variational Ordinal GP AEs (\vgStef)}
\label{vogpae}
We employ the \vgStef \cite{eleftheriadis2016variational} approach to model the second VAE in \dcod: $\vect{Z}_0 \in \mathbb{R}^{N_{D} \times N_{D_0}}$ being  the input and $\vect{Z}_1$ the corresponding latent variables. Similar to VC-AEs (Sec.~(\ref{vcae})),  the objective of this layer becomes:
\begin{align}\label{final_elbo}
\begin{split}
\mathcal{L}_{\vgStef}(W_o, \theta_{GP}, V) &= \mathcal{L}_{kl, \vect{Z}_0} + \mathcal{L}_{r, \vect{Z}_0} + \mathcal{L}_{o, \vect{Z}_0} \\
\mathcal{L}_{kl, \vect{Z}_0} &= - D_{KL}(q(\vect{Z}_1|\vect{Z}_0)||p(\vect{Z}_1)) \\
\mathcal{L}_{r, \vect{Z}_0} &= \sum_{d=0}^{D_0} \mathbb{E}_{q(\vect{Z}_1|\vect{Z}_0)} \big[ \log p(\vect{z}_0^d| \vect{Z}_1)\big]\\
\mathcal{L}_{o, \vect{Z}_0} &= \mathbb{E}_{q(\vect{Z}_1|\vect{Z}_0)} [\log p(\vect{Y} | \vect{Z_1}, W_o)],
\end{split}
\end{align}
where 
\begin{align}\label{ae}
 \vect{Z}_0|\vect{Z}_{1} &= f_{\vect{Z}_{1} \to \vect{Z}_0}(\cdot; \theta_{\vect{Z}_{1} \to \vect{Z}_0}), \\ 
 \vect{Z}_{1}| \vect{Z}_0 &= f_{\vect{Z}_{0} \to \vect{Z}_1}(\cdot;\theta_{\vect{Z}_{0} \to \vect{Z}_1}), \\ 
 \vect{Y}|\vect{Z}_{1} &=  f_{\vect{Z}_{1} \to \vect{Y}}(\cdot; W_o).
\end{align}
Here, $f_{\vect{Z}_{1} \to \vect{Z}_0}$ and $f_{\vect{Z}_{0} \to \vect{Z}_1}$ are the encoding and decoding mappings with GP priors, $\theta_{\vect{Z}_{1} \to \vect{Z}_0}$ and $\theta_{\vect{Z}_{0} \to \vect{Z}_1}$ are the corresponding kernel parameters, and $f_{\vect{Z}_{1} \to \vect{Y}}$ is the classification function. We place GP priors on both mappings, resulting in:
\begin{align}
 p(\vect{Z_0}|\vect{Z_1}) = \mathcal{N}(\vect{0}, \vect{K}_{\vect{Z}_1 \to \vect{Z}_0} + \sigma_v^2\vect{I}),  \\  
 p(\vect{Z_1}|\vect{Z_0}) = \mathcal{N}(\vect{0}, \vect{K}_{\vect{Z}_0 \to \vect{Z}_1} + \sigma_r^2\vect{I}),
\end{align}

\begin{align}\label{likelihood}
p(\vect{Z}_0) = \int \prod_{d=1}^{D_0} p(\vect{z}_0^d|\vect{Z}_1)p(\vect{Z}_1)d\vect{Z}_1,
\end{align}
where $D_o$ is the dimension of $\vect{Z}_0$. Since computing its marginal likelihood is intractable (due to the non-linear coupling of the GP kernels), we resort to approximations. To  this end, the approximate variational distribution $q(\vect{Z_1}|\vect{Z_0})$ is used to recover a Bayesian non-parametric solution for both the GP encoder \& decoder, and is defined as:
\begin{align}\label{var_dist}
q(\vect{Z}_1|\vect{Z}_0) = \prod\nolimits_i \mathcal{N}(\hat{\vect{m}}_i, \vect{S}_i + \hat{\sigma}_i^2I),
\end{align}
where $\vect{M}=\{\vect{m}_i\}, \; i=1,...,N$ and $\vect{S}=\{\vect{S}_i\}, \; i=1,...,N$ are variational parameters, and $\hat{\vect{m}}_i = \vect{m}_i - \left[\vect{K}_{\vect{Z}_1 \to \vect{Z}_0}^{-1}\vect{M} \right]_i / \left[\vect{K}_{\vect{Z}_1 \to \vect{Z}_0}^{-1} \right]_{ii}$ and $\hat{\sigma}_i^2 = 1 / \left[\vect{K}_{\vect{Z}_1 \to \vect{Z}_0}^{-1} \right]_{ii}$ is the leave-one-out solution of GP \cite{rasmussen2006gaussian}. 

We further constrain the latent variable $\vect{Z_1}$ by imposing the ordinal structure on the output labels $\vect{Y}$ as:
\begin{align}\label{ord_noisy}
p(\vect{Y}|\vect{Z_1}) &= \prod_{i,c} \mathbbm{I}{( \vect{y}_i=c )} p(\vect{y}_{i}|\vect{z}_{1i}) \quad , \\
p(\vect{y}_{i} = s|\vect{z}_{1i}) &= 
\begin{cases}
1 \quad $if $ f_{\vect{Z}_{1} \to \vect{Y}}(\vect{z}_{1i}) \in (\gamma_{c,s-1}, \gamma_{c,s}]\\
0 \quad $otherwise$,
\end{cases},\\
f_{\vect{Z}_{1} \to \vect{Y}}(\vect{z_{1i}}) &= w_o^T \vect{z}_{1i} + \epsilon, \; \epsilon \sim \mathcal{N}(0, \sigma^2_o),
\end{align}
where $\mathbbm{I}{(\mathord{\cdot})}$ is the indicator function that returns 1 (0) if the argument is true
(false) and $i=1,\dots,N$ indexes the training data. $\gamma_{c,0} = -\infty \le \cdots \le \gamma_{c,S} = +\infty$ are the thresholds or cut-off points that partition the real line into $s=1,\dots,S$ contiguous intervals. We arrive at the ordinal log-likelihood (see~\cite{chu2005gaussian} for details):
\begin{equation}\label{ord_lklhd}
    \begin{split}
    \mathbb{E}_{q(\vect{Z}_1|\vect{Z}_0)} (\log p(\vect{Y} | \vect{Z}_1, \vect{W}_o)) = \sum_{i,c}\mathbb{I}(y_{ic}=s) \log\\ \biggl(\Phi\left(\frac{\gamma_{c,s} - \vect{w}_o^T\vect{z}_{1i}}{\sigma_o}\right)  - 
\Phi\left(\frac{\gamma_{c,s-1} - \vect{w}_o^T\vect{z}_{1i}}{\sigma_o}\right)\biggr)
    \end{split}
\end{equation}
where $\Phi(\cdot)$ is the Gaussian cumulative density function.

The random process of recovering the latent variables has two distinctive stages: (a) the latent variables $\vect{Z}_{1}$ are generated from some general prior distribution $p(\vect{Z_1})=\mathcal{N}(\vect{0}, \vect{I})$, and further projected to the labels' ordinal plane via $p(\vect{Y}|\vect{Z_1})$; (b) the input $\vect{Z}_1$ is generated from the conditional distribution $p(\vect{Z}_1|\vect{Z}_0)$. The model parameters are: $\theta_{GP} = \{\theta_{\vect{Z}_{1} \to \vect{Z}_0}, \;\theta_{\vect{Z}_{0} \to \vect{Z}_1}\}$, $W_o = \{w_o, \;\sigma_o\}$, and $V=\{M,S\}$ are variational parameters. 


\section{Learning and Inference}

Learning of \dcod consists of maximizing the joint lower bound (Sec.~(\ref{sub:jointlearning})) w.r.t the VC-AEs parameters ($W_d$, $\mu$, $\sigma$) and the VO-GPAE (hyper-) parameters ($V$, $\theta_{GP}$,$W_o$).\footnote{This is not an exact lower bound for target objective function but a combination of the two bounds obtained via coupling of the posteriors.} For the GP-encoder/decoder kernel, we use the radial basis function (RBF) with automatic relevance determination (ARD), which can effectively estimate the dimensionality of the latent space \cite{damianou2012manifold}. For both VC-AE and VO-GPAE, we utilize a joint optimization scheme using stochastic backpropagation \cite{rezende2014stochastic}, with the re-parameterization trick \cite{kingma2013auto}. Before we detail the steps of our learning algorithm, we first describe the proposed iterative balanced batch learning (Sec.~(\ref{sub:balancedbatch})) and the warming criterion to efficiently learn the latent features (Sec.~(\ref{sub:warming})). These strategies turn out to be critical in avoiding overfitting and achieving significant learning speed-ups.

\subsection{Iterative Balanced Batch Learning}\label{sub:balancedbatch}
Minimizing the model objective using all training data can easily lead to a local minimum, and, thus, poor performance. This is due to the inherent hierarchical structure of the model (VAE layers), and highly imbalanced AU intensity labels. We introduce an iterative balanced batch learning approach to deal with the data imbalance. The main idea is to update each set of parameters with batches that are balanced with respect to subjects in the dataset (number of example images of each subject) and AU intensity levels. This ensures that the network used for facial feature extraction is not biased towards a specific subject/AU level. We use Stochastic Gradient Descent (SGD) with a batch size of 32, learning rate of 0.01 and momentum of 0.9.

\subsection{Warming Strategy}\label{sub:warming}
The lower bound in Eq.~(\ref{eq:lossCVAE}\&\ref{final_elbo}) consists of three terms. Each model that encodes a latent variable $\vect{Z}_i$ will have a non-zero KL term and a relatively small cross-entropy term. Practically, implementations of such AEs will struggle to learn this behavior. As pointed out in \cite{sonderby2016ladder, bowman2015generating, raiko2007building}, training these models will lead to consistently setting the approximate distribution $q(\vect{Z}_i|\vect{Z}_{i-1})$ equal to the prior $p(\vect{Z}_i)$, and thus bringing the KL divergence to zero. This can be of advantage and seen as ARD, but also be a challenge in training for the latent space to learn a useful (and discriminative) representation. To avoid this, we propose different warm-up strategies for training VAEs in \dcod. Specifically, instead of directly maximizing the lower bound of the VC-AE (Eq.(\ref{eq:lossCVAE})), we augment the learning by including the expectation of the predicted labels ($\mathcal{L}_{p}$) for intensity classification of AUs, steering the parameters towards more discriminative latent representations. Formally, this is attained by using the weighted objective given by:
\begin{align}
\mathcal{L}_{VC-AE}=\alpha \mathcal{L}_{kl, \vect{X}} + \mathcal{L}_{r, \vect{X}} + (1-\alpha)\mathcal{L}_{p, \vect{X}},
\end{align}
where 
\begin{align}
\mathcal{L}_{p, \vect{X}} &= \mathbb{E}_{q(\vect{Z}_0|\vect{X})} \big[\log(p(\vect{Y}|\vect{Z}_0, W_d))\big], \\
\vect{Y}|\vect{Z}_{0} &=  f_{\vect{Z}_{0} \to \vect{Y}}(\cdot; W_c).
\end{align}
Here, $(\vect{Y}|\vect{Z_0})$ can be modeled using any classifier $W_c$ (we used logistic regression). Note that initially ($\alpha=0$) we focus on finding a discriminative subspace at the first layer of \dcod. With the  increasing number of iterations, the KL divergence term overtakes the classification loss, assuring the smoothness of the subspace $\vect{Z}_{0}$. We then construct a lower bound for \vgStef with a warming term as:
\begin{align}
\mathcal{L}_{VO-GPAE} &= \beta \mathcal{L}_{kl, \vect{Z}_0} + \mathcal{L}_{r, \vect{Z}_0} + \mathcal{L}_{o, \vect{Z}_0}.
\end{align}
Both $\alpha$ and $\beta$ are linearly increased from 0 to 1 during the first $N_t$ epochs of training. Note that in the beginning, we include the classification loss in the first layer - which acts as a regularizer. However, it slowly diminishes as we obtain more stable estimates of the variational distributions at each layer, since toward reaching the $N_t$-th epoch, the VO-GPAE classifier stabilizes and $\vect{Z}_0$ need no more be class-regularized. We found that this approach works very well in practice, as shown in Sec.~(\ref{experiments}).

\subsection{Joint Learning}\label{sub:jointlearning}
In the 2-layer \dcod, we optimize the lower bound:
\begin{align}\label{eq:deepcoderelbo}
\mathcal{L}_{DC} = \mathcal{L}_{VC-AE} + \mathcal{L}_{VO-GPAE}.
\end{align}
The main bottleneck of the second AE is that it cannot use all training data as the computation of covariance function in \vgStef would be prohibitively expensive. Because of this, we propose a 'leave-subset-out' strategy, where we learn the target AEs in an iterative manner. Specifically, we split the training dataset $\vect{X}$ in two non-overlapping subsets, $X_R$ and $X_L$, $X_R>>X_L$. $X_R$ is used for training VC-AE, while $X_L$ is used for training \vgStef. First, VC-AE is initialized using $X_R$ by minimizing Eq.~(\ref{eq:lossCVAE}) for 5 epochs, followed by the two-step iterative training algorithm. In the first step, we find the latent projections using $X_L$, i.e., $\vect{Z}_{0,L}$ by VC-AE and learned parameter $W_d$, $\mu_R$ and $\sigma_R$ from $X_R$. $\vect{Z}_{0,L}$ are then used to train VO-GPAE for one epoch, minimizing Eq.~(\ref{final_elbo}). In the second step, we reconstruct $X_R$ as $\vect{Z}_{0,R}$, and also compute the posteriors  $\tilde{p}(\vect{Z}_{0,R})$, which are then fed into the VC-AE to update its parameters by minimizing Eq.~(\ref{eq:lossCVAE}) in one epoch. These two steps are repeated until convergence of the joint lower bound $\mathcal{L}_{2DC}$. In this way, we constantly update the prior on $\vect{Z}_0$, which propagates the information from the bottom VO-GPAE, effectively tying the parameters of the two AEs.
\begin{algorithm}[t]
\small
\caption{\dcod: Learning and Inference}
\label{alg}
\begin{algorithmic}
\\\hrulefill
\State \textbf{Learning}: Input $\mathcal{D}_{tr}=(\vect{X}, \vect{y})$
\State Split $\vect{X}\in\vect{X}_R \cup \vect{X}_L$, $\vect{X}_R>>\vect{X}_L$, and $ \vect{X}_R \cap \vect{X}_L = \emptyset$.
\Repeat
\State \textbf{if} \text{init run,} $p(\vect{Z}_{0,R}) \sim \mathcal{N}(0, 1)$ 
\State \textbf{else}  $\tilde{p}(\vect{Z}_{0,R}) = p(\vect{Z}_{0,R}|\vect{Z}_{1,L})$ \textbf{end}
\State \textbf{Step 1:} for 1 epoch, optimize $\mathcal{L}_{VC-AE}$ given  $\vect{X}_R$, 
\State  \,\,\,\,\,\,$\vect{Z}_{0, R} = f_{\vect{X} \to \vect{Z_0}}(X_R)$ \, and \, $\vect{Z}_{0, L} = f_{\vect{X} \to \vect{Z_0}}(X_L)$
\State \textbf{Step 2:} for 1 epoch, optimize $\mathcal{L}_{VO-GPAE}$ given $\vect{Z}_{0, L}$,
\State  \,\,\,\,\,\,$\vect{Z}_{0, R} = f_{\vect{Z_1} \to \vect{Z_0}}(f_{\vect{Z}_0 \to \vect{Z}_1}(Z_{0, R}))$
\Until{convergence of $\mathcal{L}_{2DC}$}
\State Output: $W_d, \mu_R, \sigma_R, W_o, \theta_{GP}, V_L$
\\\hrulefill
\State \textbf{Inference}: \text{Input} $\mathcal{D}_{te}=(\vect{X}_{*})$
\State \textbf{Step 1:} $\vect{Z}_{1, *} = f_{\vect{Z}_0 \to \vect{Z}_1}(f_{\vect{X} \to \vect{Z}_0}(\vect{X}_{*}, W_d))$
\State \textbf{Step 2:} $\vect{y}_{*} = f_{\vect{Z_1} \to \vect{Y}}(\vect{Z}_{1, *},W_o)$
\State  \,\,\,\,\,\,\,\,\,\,\,\,\,\,\,\,\,\,$\vect{X}'_{*} = f_{\vect{Z}_0 \to \vect{X}}(f_{\vect{Z}_1 \to \vect{Z}_0}(\vect{Z}_{1, *}), W_d)$
\State \text{Output:} $\vect{X}'_{*}, \; \vect{y_{*}}$
\end{algorithmic}
\end{algorithm}
\begin{table*}[t]
\setlength{\tabcolsep}{1.0pt}  
\center
\caption{Performance of different models for AU intensity estimation on the DISFA and FERA2015 database. DC- and CNN-based models were trained using raw images as input. The results for the models highlighted with $\dagger$ were taken from \cite{eleftheriadis2016variational} (the model trained with LBP+landmark features). The model highlighted with $\star$ was trained with the deep features, extracted from the last layer of the best performing \cnn, and, thus, is directly comparable to the proposed \textbf{\dc}.}
\scalebox{0.8}{
\begin{tabularx}{\textwidth}{|ll|XXXXXXXXXXXX|X|XXXXX|X|}
    \hline
    \rowcolor{Gray}
        \hline
\multicolumn{2}{|l|}{\quad Dataset:} & \multicolumn{13}{c|}{{DISFA}}           & \multicolumn{6}{c|}{{FERA2015}} \\
	\rowcolor{Gray}
\multicolumn{2}{|l|}{\quad AU:} & 1 & 2    & 4    & 5    & 6   & 9   & 12   & 15  & 17  & 20  & 25  & 26   & Avg. & 6   & 10   & 12   & 14   & 17   & Avg. \\
    \hline
    \toprule
    \hline
    \parbox[t]{3mm}{\multirow{13}{*}{\rotatebox[origin=c]{90}{\qquad ICC}}}
& \textbf{\dc} & \textbf{.70}  & \textbf{.55}  & \textbf{.69}  & .05 & .59 & \textbf{.57}  & \textbf{.88} & .32 & .10 & .08 & \textbf{.90}  & .50  & \textbf{.50}  & \textbf{.76}  & .71  & .85  & .45  & \textbf{.53}   & \textbf{.66}  \\
& \dccnn & .52  & .49  & .48  & .18 & .59 & .39  & .74 & .15 & .26 & .08 & .80  & .44  & .43 & .74  & \textbf{.72}  & .84  & .33  & .52   & .63  \\
& \cnn & .58  & .52  & .55  & \textbf{.20} & .59 & .42  & .78 & .08 & .25 & .04 & .84  & \textbf{.54}  & .44 & .76  & .70  & .85  & .36  & .49   & .63  \\
& \ocnn & .33  & .31  & .32  & .16 & .32 & .28  & .71 & .33 & \textbf{.44} & .27 & .51  & .36  & .36 & .71  & .63  & .87  & .41  & .31   & .58  \\
& \drcnn & .18  &.15   &.61   &.07  & \textbf{.65} &.55   &.82  &\textbf{.44}   &.37    &\textbf{.28}   &.77   &\textbf{.54}   &.45  &.75   &.69   &.86   &.40   &.45    & .63\\
& \vgg & .46  & .44  & .44  & .06 & .44 & .34  & .59 & .01 & .11 & .03 & .71  & .42  & .32 & .63  & .61  & .73  & .25  & .31   & .51  \\
& \vgStef \cite{eleftheriadis2016variational}$^\star$  & .18  & .00  & .27  & .15 & .57 & .34  & .80 & .01 & .00 & .02 & .88  & .55  & .31 & .72  & .66  & .78  & .43  & .56   & .63  \\
& \vgStef \cite{eleftheriadis2016variational}$^{\dagger}$ &.48 &.47 &.62 &.19 &.50 &.42 &.80 &.19 &.36 &.15 &.84 &.53 &.46 &.75 &.66 &\textbf{.88} &\textbf{.47} &.49 &.65 \\
& VAE-DGP~\cite{dai2015variational}$^\star$  & .37  & .32  & .43  & .17 & .45 & .52  & .76 & .04 & .21 & .08 & .80  & .51  & .39 & .70  & .68  & .78  & .43  & .31   & .58  \\
& GP~\cite{rasmussen2006gaussian}$^\star$ & .26  & .11  & .32  & .12 & .45 & .32  & .31 & .02 & .18 & .06 & .85  & .42  & .28 & .61  & .57  & .71  & .32  & .35   & .51  \\
& SOR~\cite{agresti2010analysis}$^\star$ & .15  & .13  & .34  & .03 & .48 & .22  & .78 & .00 & .10 & .06 & .79  & .42  & .29 & .61  & .57  & .77  & .29  & .27   & .50  \\
& MLR$^\star$ & .45  & .39  & .30  & .11 & .52 & .26  & .72 & .09 & .00 & .01 & .82  & .39  & .29 & .74  & .67  & .81  & .42  & .25   & .57  \\
    \hline
    \hline
\parbox[t]{3mm}{\multirow{13}{*}{\rotatebox[origin=c]{90}{\qquad MSE}}}  
& \textbf{\dc} & \textbf{.32}  & .39  & \textbf{.53}  & .26 & .43 & .30  &.25 & .27 & .61 & .18 & .37  & .55  & \textbf{.37} & \textbf{.75}  & 1.02 & \textbf{.66}  & 1.44  & .88   & \textbf{.95} \\
& \dccnn & .35  & .44  & .90  & \textbf{.03} & .36 & .36  & .37 & .26 & .30 & .19 & .71  & .57  & .40 & .85  & 1.03 & .75  & 1.80 & 0.81  & 1.05 \\
& \cnn   & .34  & .39  & .81  & .05 & .37 & .38  & .34 & .27 & .31 & .24 & .63  & .49  & .38 & .80  & 1.06 & .66  & 1.57 & .96   & 1.01 \\
& \ocnn & .41  & .44  & .91  & .12 & .42 & .33  & .31 & .42 & .35 & .27 & .71  & .51  & .43 & .88  & 1.12 & .68  & 1.52 & .93   & 1.02 \\
& \drcnn &.76   & .40   & .74  & .07  &.54 &.41 &.33 &\textbf{.14} &.33 &.20  &.66  &\textbf{.41} &.41 &1.23 &1.69  &.98  &2.72 &1.17 &1.57\\
& \vgg & .41  & .54  & 1.14 & .07 & .39 & .47  & .40 & .29 & .53 & .19 & .64  & .51  & .39 & .93  & 1.04 & .91  & 1.51 & 1.10  & 1.10  \\
& \vgStef \cite{eleftheriadis2016variational}$^\star$   & 1.18 & .77  & 1.14 & .11 & \textbf{.22} & .53  & \textbf{.16} & .18 & .99 & .81 & \textbf{.21}  & .46  & .56 & 0.9  & \textbf{.98}  & .67  & 1.81 & 1.31  & 1.11 \\
& \vgStef \cite{eleftheriadis2016variational}$^{\dagger}$ &.51 &\textbf{.32} &1.13 &.08 &.56 &.31 &.47 &.20 &.28 &.16 &.49 &.44 &.41 &.82 &1.28 &.70 &1.43 &\textbf{.77} &1.00\\
& VAE-DGP~\cite{dai2015variational}$^\star$  & 1.02 & 1.13 & .92  & .10 & .67 & \textbf{.19}  & .33 & .46 & .58 & .19 & .69  & .65  & .57 & .93  & 1.15 & .80  & 1.66 & 1.14  & 1.13 \\
& GP~\cite{rasmussen2006gaussian}$^\star$ & .49  & .60  & 1.06 & .08 & .38 & .30  & .26 & .25 & .30 & .19 & .61  & .69  & .63 & 1.07 & 1.27 & 1.03 & 1.52 & 0.94  & 1.17 \\
& SOR~\cite{agresti2010analysis}$^\star$ & 1.35 & .57  & 1.43 & .09 & .46 & 1.48 & .40 & .25 & .62 & .49 & 1.27 & .93  & .78 & 1.59 & 1.71 & 1.06 & 2.90 & 2.24  & 1.90 \\
& MLR$^\star$ & .42  & .49  & 1.04 & .05 & .40 & .33  & .45 & .23 & \textbf{.24} & \textbf{.13} & .62  & .55  & .41 & .84  & 1.06 & .72  & \textbf{1.35} & 1.04  & 1.00 \\
    \hline
\end{tabularx}
}
\label{tab:results}
\end{table*} 

Inference in the proposed \dc: the test data $\vect{X}_*$ is first projected to the latent space $\vect{Z}_0$ via the VC-AE, and then further passed through the VO-GPAE via $\vect{Z}_1$. The obtained latent positions are then used for ordinal classification of target AU intensities. The decoding starts with the reconstruction of the latent points in $\vect{Z}_0$, followed by the reconstruction of $\vect{X}_*$. These steps are summarized in Alg.~(\ref{alg}).

\section{Experiments}\label{experiments}
{\bf Datasets.}
We evaluate the proposed \dcod on two benchmark datasets for AU intensity estimation: DISFA~\cite{mavadati2013disfa} and FERA2015 challenge data~\cite{valstar2015fera}. Both contain per frame AU intensity annotations on a 6-point ordinal scale (DISFA 12, FERA2015 6 AUs). Also, we performed subject-independent validation: DISFA (3 folds: 18 train/9 test subjects), and FERA2015 (2 fold: 21 train / 20 test).

\textbf{Pre-Processing.} 
For the CNN-based models, we used the dlib face detector \cite{kazemi2014one} to extract the face location from images in each dataset. We then registered the 49 facial points to a reference frame (average points in each dataset) using a similarity transform and cropped a bounding box of $240 \times 160$ pixel size. These were then normalized using per-image histogram-equalization, which increases the robustness against illumination changes. For models in which it is not feasible to process high dimensional features from raw images, we extracted the 2000-D features ($\vect{Z}_0$) from the CNN - in our experiments, this size was found optimal for the competing methods. During evaluation, we used the negative log-predictive density (NLPD) for the reconstruction error, and for classification the mean squared error (MSE), the classifier's consistency of the relative order of the intensity levels, and intra-class correlation (ICC(3,1))~\cite{shrout1979intraclass} - agreement between annotators.  
\\
\\
\textbf{Models.} 
As a baseline, we use the multivariate linear regression (MLR) for joint estimation of AU intensities and the standard ordinal regression (SOR)~\cite{agresti2010analysis} serves as the second baseline. The \cnn model is a standard 2-layer CNN for multi-output classification (we used the same setting as in \cite{gudi2015deep}). The \ocnn is an ordinal CNN that was originally introduced for the task of age estimation; we applied it to our task. \vgg is a widely used NN for object detection. To adapt it for our task, we used the pre-trained model and fine-tuned the last 3 layers. As a baseline for the GP-based models, we use the standard GP~\cite{rasmussen2006gaussian} with a shared covariance function among outputs. We also compare the proposed to \vgStef \cite{eleftheriadis2016variational}, the state-of-the-art GP model for variational ordinal regression. Here, we evaluated the model on two sets of features: LBPs with facial landmarks, and deep features, extracted using the CNN (our first coder). We evaluate the proposed model in two settings: \dccnn is the fully parametric \dcod (\dccnn), where we simply apply a stack of two VC-AEs with a 50D latent space ($\vect{Z}_1$) and 2000D ($\vect{Z}_0$) features- as also set in our semi-parametric \dc model, with \vgStef at the bottom layer. For the iterative {\bf 2DC} learning algorithm, we split the dataset according to the algorithm in two subsets $N_L$ and $N_R$. Due to the computational complexity of GPs ($\mathbb{O}(N^3)$), we chose a rather small subset of $N_L = 5000$ to train the VO-GPAE, while using the rest of data set for our convolutional auto-encoder VC-AE ($N_R = 71223$ for FERA2015 and $N_R=87209$ for DISFA). For subset $N_L$, we also chose a subject balanced subset, i.e. every subject is equally represented in the number of frames. We used the pre-processed raw images as input to the proposed \dcod and CNN based models. As the GP-based models and the other baselines are not directly applicable to high dimensional image data, we trained on the LBP+landmark features and/or deep features, extracted from the last layer of the best performing \cnn model. For the sake of comparisons, we also include the results from the recently published deep structured learning model with the database augmentation - \drcnn (we show the reported results).

\section{Results}
\begin{figure}
    \begin{subfigure}{.5\linewidth}
        \centering
        \includegraphics[width=1\linewidth]{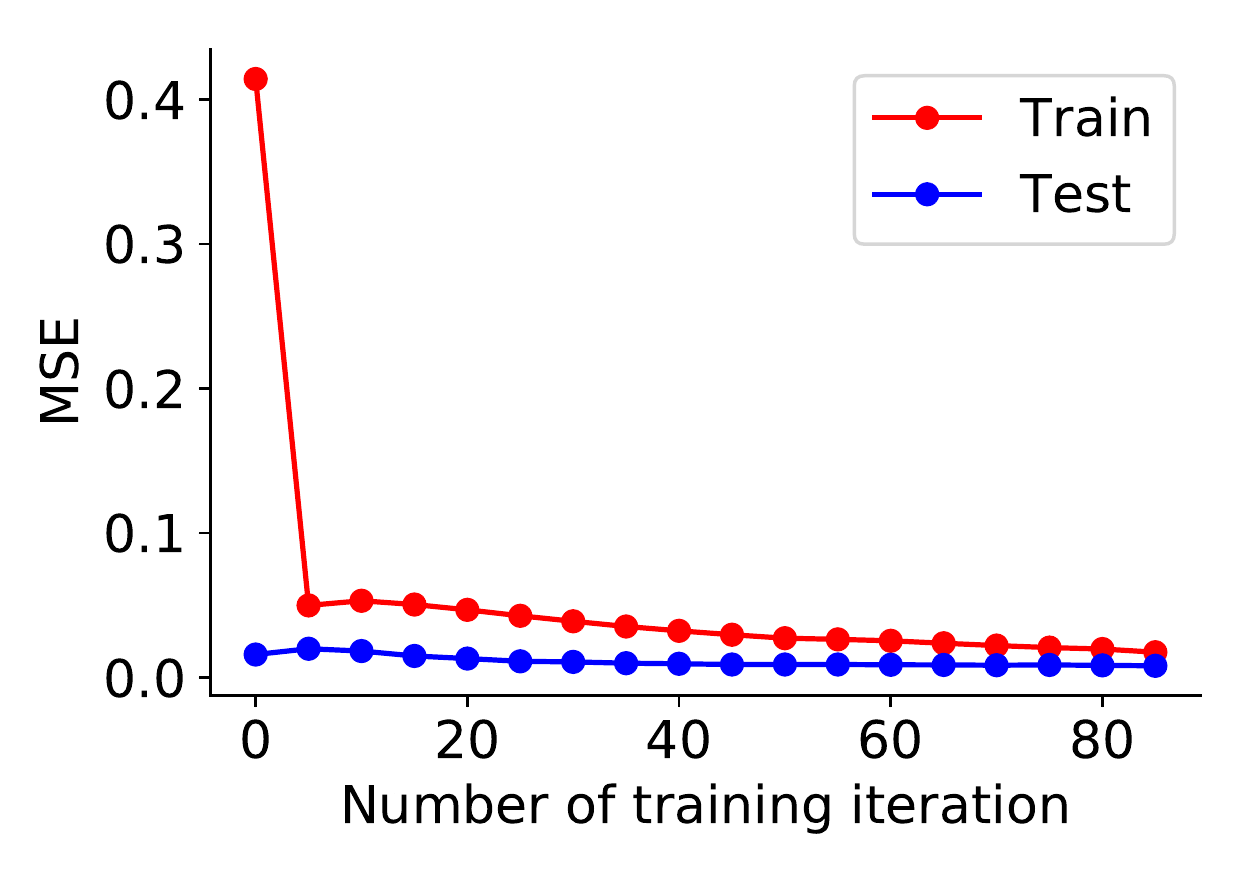}
        \caption{MSE}
        \label{fig:mse}
    \end{subfigure}%
    \hfill
    \begin{subfigure}{.5\linewidth}
        \centering
        \includegraphics[width=1\linewidth]{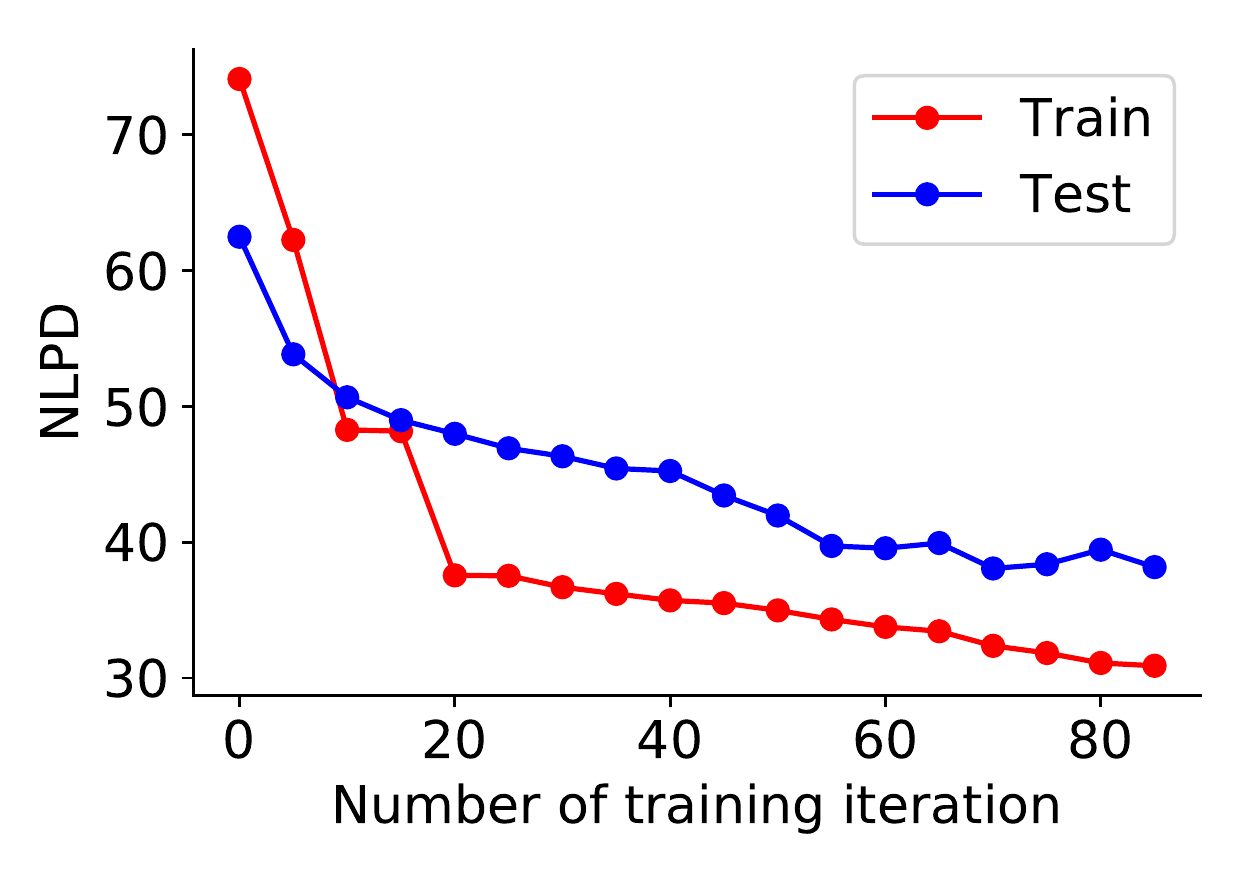}
        \caption{NLPD}
        \label{fig:nlpd}
    \end{subfigure}
    \begin{subfigure}{.5\linewidth}
        \centering
        \includegraphics[width=1\linewidth]{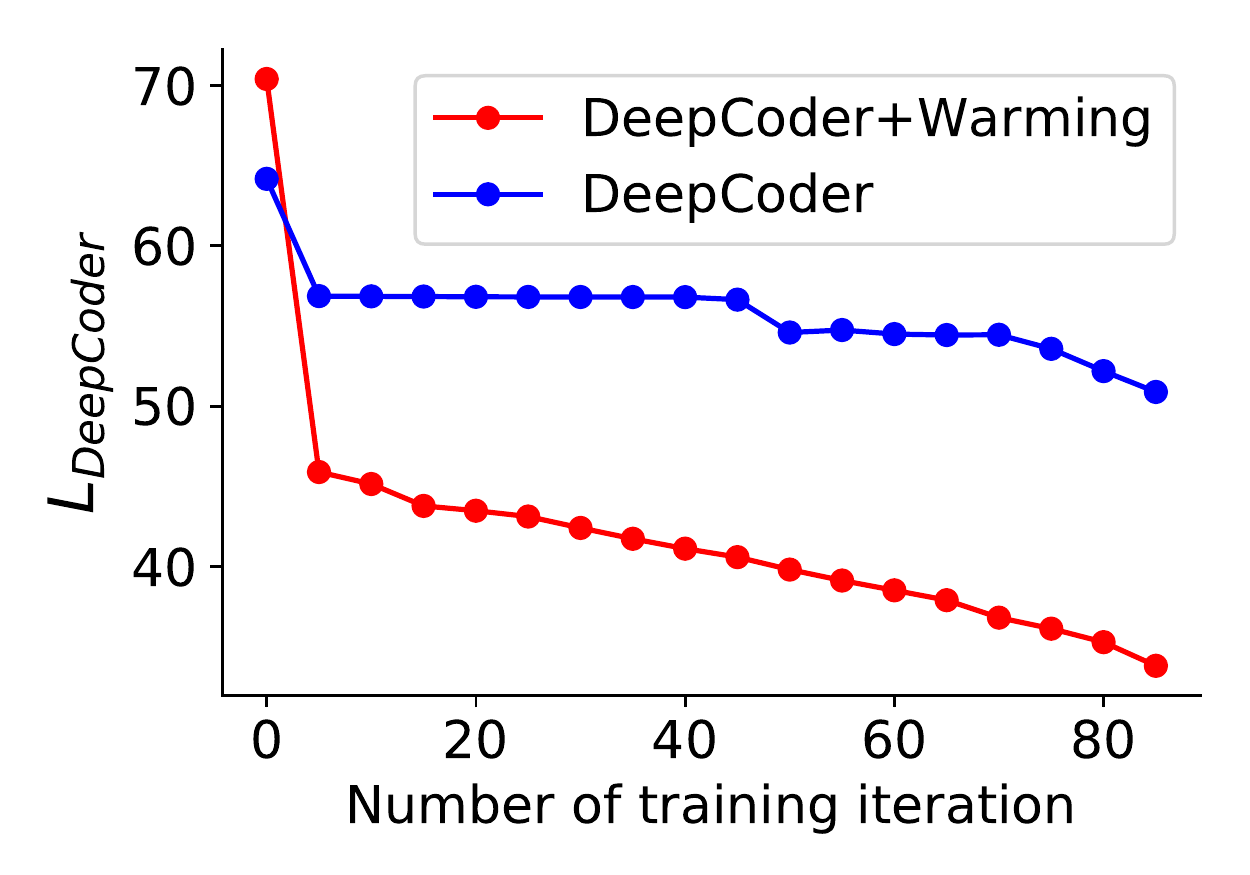}
        \caption{Lower bounds}
        \label{fig:loss}
    \end{subfigure}%
    \hfill
    \begin{subfigure}{.5\linewidth}
        \centering
        \includegraphics[width=1\linewidth]{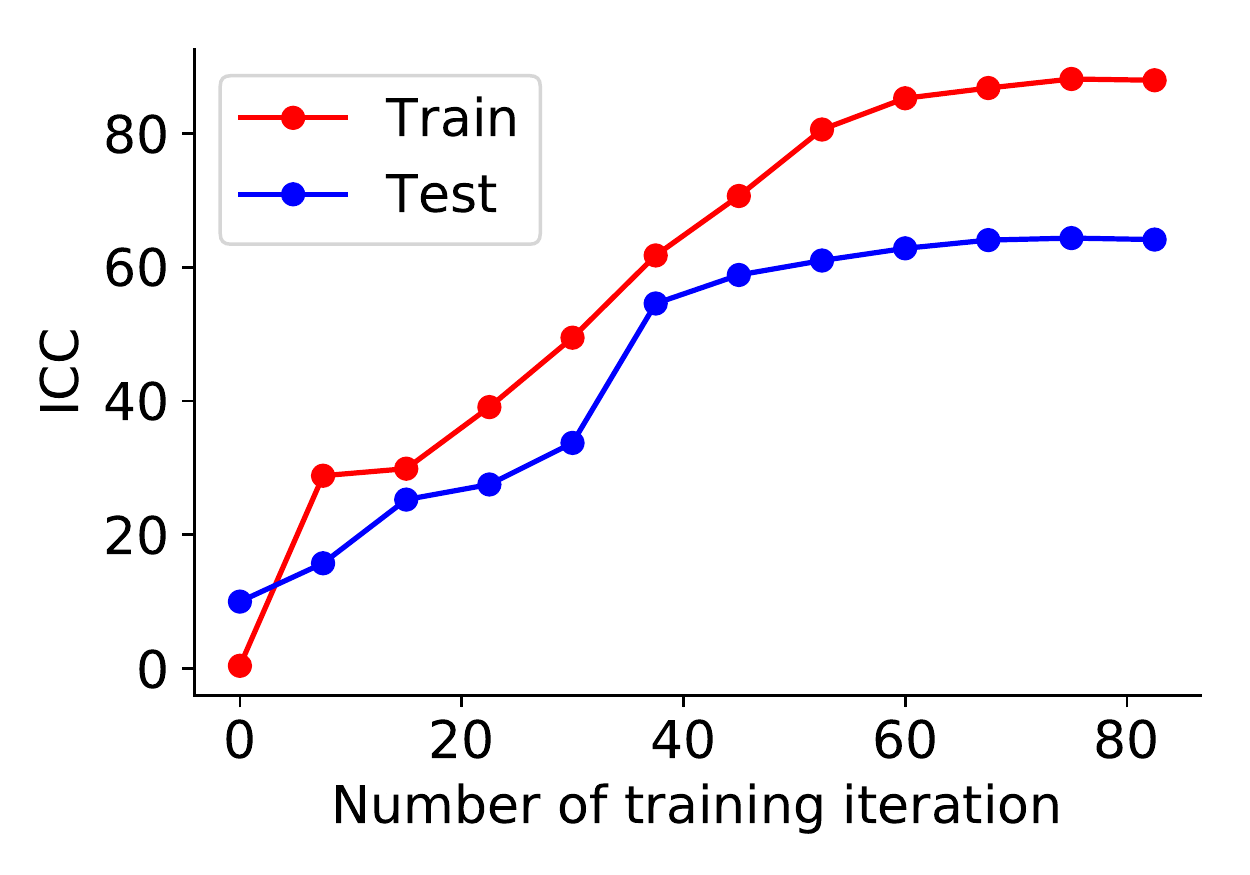}
        \caption{ICC}
        \label{fig:icc}
    \end{subfigure}
    \begin{subfigure}{.5\linewidth}
        \centering
        \includegraphics[width=1\linewidth]{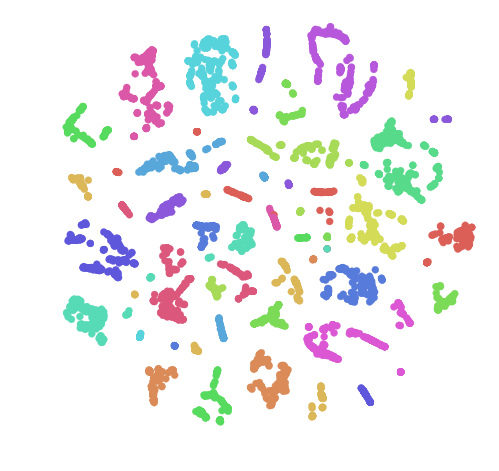}
        \caption{Latent space $\vect{Z}_0$}
        \label{fig:feat}
    \end{subfigure}%
    \hfill
    \begin{subfigure}{.5\linewidth}
        \centering
        \includegraphics[width=1\linewidth]{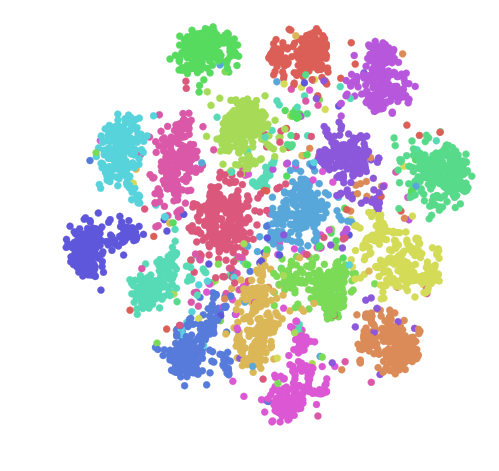}
        \caption{Latent space $\vect{Z}_1$}
        \label{fig:mu}
    \end{subfigure}
        \caption{FERA2015: (a) the MSE reconstruction error, (b) the NLPD of VO-GPAE, (c) the estimated variational lower bound per data point, (d) ICC for the AU intensity estimation, and the recovered latent spaces: $\vect{Z}_0$ (e), and $\vect{Z}_1$ (f).}
\label{fig:plots}
\end{figure}

\textbf{Quantitative Results.} Table~(\ref{tab:results}) shows the comparative results. On average, the CNN based models largely outperform the GPs in both measures across most of the AUs. This is because CNNs are capable to jointly learn the embedded space and classifier from raw images, while GPs are trained on hand-crafted features, which turn out to be less discriminative for the task. This can be particularly observed from AU17 in both datasets. Also, both the relative shallow \cnn and the \dccnn model achieve an ICC of 44\%/43\% on DISFA and 63\% on FERA2015, which is highest performance using current deep models. By comparing the predictions of these two models, we see that the performance of the fully parametric \dccnn does not increase by blindly stacking VC-AEs on top of each other. The same applies to the basic CNNs models. Furthermore, both models are outperformed by the proposed semi-parametric \dc. This is mainly because GPs are known to provide a better classifier (non-parametric, hence they are more flexible in modeling complex distributions). This can be seen from Fig.~( \ref{fig:plots}), where the samples on the latent space ${\bf Z}_1$ are clustered into different subjects. Note that this subject clustering in the latent space has been done in an unsupervised manner by GPs (i.e., no subject id was provided). The bottom VO-GPAE layer benefits from the robust features coming from the top VC-AE, and the jointly learned ordinal classifier using the proposed iterative algorithm (Alg.~(\ref{alg})).

The standard \vgg network does not achieve competitive results with the proposed model, most likely because it does not account for ordinal intensity levels and does not perform simultaneous learning of latent features. The \ocnn model, which has the same architecture as \cnn but with the ordinal classifier, learns one binary classifier for each intensity level of each AU, resulting in a large number of parameters, easily prone to overfitting. Overall, from average results on both datasets, we clearly see the benefits of the joint learning in the proposed \dcod ({\bf 2DC}). Finally, note that the proposed \dcod outperforms the state-of-the-art approach (\drcnn), which takes advantage of CNNs and data augmentation based on multiple face datasets. Again, we attribute this to the lack of non-parametric feature learning and ordinal classifier in the latter.

\textbf{Qualitative Results.}
Fig.~(\ref{fig:plots}) shows a summary of the model loss per iteration and the learned latent spaces for the two levels of the proposed \dcod, for the FERA2015 dataset. Fig.~(\ref{fig:mse}) depicts the reconstruction error of the input images $X$ measured by (MSE) while Fig.~(\ref{fig:nlpd}) visualizes the NLPD of the latent space $\vect{Z}_0$. While the reconstruction loss of the images converges quickly after five iteration, the NLPD of $\vect{Z}_0$ steadily decreases but needs 50 iterations to converge. The reason is the initialization. The weights for the latent space $\vect{Z}_0$ were initialized according to \cite{glorot2010understanding} which has proven to converge quickly, while \vgStef was initialized by drawing randomly from a normal distribution. Thus, $\vect{Z}_1$ required more iterations to converge. In Fig.~(\ref{fig:loss}), we compare the lower bounds with and without warming strategy (see Sec.~(\ref{sub:warming})). As expected, without warming strategy, the lower bound gets stuck in a local minimum, while the warming strategy lead to a steady decrease in the bound value. From the latent spaces $\vect{Z}_0$ and $\vect{Z}_1$ in Fig.~(\ref{fig:feat},\ref{fig:mu}), we observe that $\vect{Z}_0$ is clustered according to subjects, but still the subjects are scattered over the latent space (showing the model's invariance to identity, as also shown in \cite{chu2017}). However, in $\vect{Z}_1$ space, the model fits each subject into a separate cluster. As evidenced by our results, this clustering of the subjects leads to more efficient features for AU intensity estimation. We attribute this to the fact that GPs do an efficient smoothing over the training subjects closest to the test subject in the learned subspace - evidencing the importance of addressing the subject differences using non-parametric models. 

\section{Conclusions}
We proposed a novel deep probabilistic framework, \dcod, for learning of deep latent representations and simultaneous classification of multiple ordinal labels. We showed in the context of face analysis that the joint learning of parametric features, followed by learning of the non-parametric latent features and target classifier, results in improved performance on the target task achieved by the proposed semi-parametric \dcod. We showed that this approach outperforms parametric deep AEs, and the state-of-the-art models for AU intensity estimation. 

\section*{Acknowledgements}
This work has been funded by the European Community Horizon 2020
under grant agreement no. 688835 (DE-ENIGMA), and the work of O. Rudovic under grant agreement no. 701236 (EngageMe - Marie Curie Individual Fellowship).

{\small
\bibliographystyle{ieee}
\bibliography{egbib}
}

\end{document}